\documentclass[a4paper,twoside]{article}

\usepackage{epsfig}
\usepackage{subcaption}
\usepackage{calc}
\usepackage{amssymb}
\usepackage{amstext}
\usepackage{amsmath}
\usepackage{amsthm}
\usepackage{multicol}
\usepackage{pslatex}
\usepackage{apalike}

\usepackage{hyperref}
\usepackage{algorithm}
\usepackage[noend]{algpseudocode}

\usepackage{SCITEPRESS}     

\renewcommand{\times}{*}

\theoremstyle{plain}
\newtheorem{defCounter}{Definition} 
\newtheorem{exCounter}{Example} 

\theoremstyle{definition}
\newtheorem{defn}[defCounter]{Definition} 
\newtheorem{exmp}[exCounter]{Example} 
\theoremstyle{plain}

\newcommand{\wmc}{\textsc{WMC}}
\newcommand{\weight}{\textsc{WEIGHT}}
\newcommand{\wmi}{\textsc{WMI}}
\newcommand{\vol}{\textsc{VOL}}
\newcommand{\vars}{\textsc{VARS}}

\begin{document}

\frenchspacing

\title{Scaling up Probabilistic Inference in Linear and Non-Linear Hybrid Domains by Leveraging Knowledge Compilation}
\author{\authorname{Anton R. Fuxjaeger\sup{1}, Vaishak Belle\sup{1,2}}
\affiliation{\sup{1}University of Edinburgh, United Kingdom}
\affiliation{\sup{2}Alan Turing Institute, United Kingdom}
\email{\{anton.fuxjaeger, vaishak\}@ed.ac.uk}}


\abstract{ 
Weighted model integration (\wmi) extends weighted model counting (\wmc) in providing a computational abstraction for probabilistic inference in mixed discrete-continuous domains. \wmc~has emerged as an assembly language for state-of-the-art reasoning in Bayesian networks, factor graphs, probabilistic programs and probabilistic databases. In this regard, \wmi~shows immense promise to be much more widely applicable, especially as many real-world applications involve attribute and feature spaces that are continuous and mixed. Nonetheless, state-of-the-art tools for \wmi~are  limited and less mature than their propositional counterparts. In this work, we propose a new implementation regime that leverages propositional knowledge compilation for scaling up inference. In particular, we use sentential decision diagrams, a tractable representation of Boolean functions, as the underlying model counting and model enumeration scheme. Our regime performs competitively to state-of-the-art \wmi~systems but is also shown to handle a specific class of non-linear constraints over non-linear potentials.
}

\onecolumn \maketitle \normalsize \setcounter{footnote}{0} \vfill

\section{\uppercase{Introduction}}
\label{sec:introduction}

\noindent Weighted model counting (\wmc) is a basic reasoning task on propositional knowledge bases. It extends the model counting task, or \#SAT, which is to count the number of satisfying assignments to a given propositional formula~\cite{biere2009handbook}.
In \wmc, one accords a weight to every model and computes the sum of the weights of all models. The weight of a model is often factorized into weights of assignments to individual variables. 
\wmc~has  emerged as an assembly language for numerous formalisms,  providing state-of-the-art probabilistic reasoning for Bayesian networks ~\cite{chavira2008probabilistic}, factor graphs ~\cite{choi2013compiling}, probabilistic programs ~\cite{fierens2015inference}, and probabilistic databases ~\cite{suciu2011probabilistic}. Exact \wmc~solvers are based on knowledge compilation ~\cite{darwiche2004new,muise2012d} or exhaustive DPLL search ~\cite{sang2005performing}.
These successes have been primarily enabled by the development of efficient data structures, e.g., arithmetic circuits (ACs), for representing Boolean theories, together with fast model enumeration strategies. In particular, the development of ACs has enabled a number of developments beyond inference, such as parameter and structure learning ~\cite{bekker2015tractable,liang2017learning,poon2011sum,kisa2014probabilistic,poon2011sum}. 
Finally, having a data structure in hand means that multiple queries can be evaluated efficiently: that is, exhaustive search need not be re-run for each query. 

However, \wmc~is limited to discrete finite-outcome distributions only, and little was understood about whether a suitable extension exists for continuous and discrete-continuous random variables until recently. The framework of  weighted model integration (\wmi) ~\cite{belle2015probabilistic} extended the usual \wmc~setting by allowing real-valued variables over symbolic weight functions, as opposed to purely numeric weights in \wmc. The key idea is to use formulas involving real-valued variables to define a hyper-rectangle or a hyper-rhombus, or in general, any arbitrary region of the event space of a continuous random variable, and use the symbolic weights to define the corresponding density function for that region.\wmc~is based on propositional SAT technology and, by extension, \wmi~is based on satisfiability modulo theories (SMT), which enable us to, for example, reason about the satisfiability of linear constraints over the reals~\cite{Barrett2009}. 
Thus, for every assignment to the Boolean and continuous variables, the \wmi~problem defines a density. The \wmi~for a knowledge base (KB) $\Delta$ is computed by integrating these densities over the domain of solutions to $\Delta$, which is a mixed discrete-continuous space, yielding the value for a probabilistic query. The approach is closely related to the mixture-of-polynomials density estimation for hybrid Bayesian networks~\cite{shenoy2011inference}. 
Applications of \wmi~(and closely related formulations) for probabilistic  graphical modelling and probabilistic programming tasks have also  been emerging ~\cite{chistikov2017approximate,albarghouthi2017quantifying,morettin2017efficient}. 

Given the popularity of \wmc, \wmi~shows immense promise to be  much more widely applicable, especially as many real-world applications, including time-series models, involve attribute and feature spaces that are  continuous and mixed. However, state-of-the-art tools for \wmi~are limited and significantly less mature than their propositional counterparts. Initial developments on \wmi~~\cite{belle2015probabilistic} were based on the so-called block-clause strategy, which naively enumerates the models of a $\mathcal{LRA}$ (\underline{l}inear \underline{r}eal \underline{a}rithmetic) theory  and is impractical on all but small problems.
Recently, a solver based on predicate abstraction was introduced by ~\cite{morettin2017efficient} with strong performance, but since no explicit circuit is constructed, it is not clear how tasks like parameter learning can be realized. Following that development, ~\cite{kolb2018efficient} proposed the use of extended algebraic decision diagrams ~\cite{sanner2012symbolic}, an extension of algebraic decision diagrams ~\cite{bahar1997algebric}, as a compilation language for \wmi. They also perform comparably to~\cite{morettin2017efficient}.
 
However, while this progress is noteworthy, there are still many significant differences to the body of work on propositional circuit languages. For example, properties such as canonicity have received considerable attention for these latter languages ~\cite{van2015role}. Many of these languages allow (weighted) model counting to be computed in time linear in the size of the obtained circuit. To take advantage of these results, in this work we revisit the problem of how to leverage  propositional circuit languages for \wmi~more carefully and develop a generic implementation regime to that end. In particular, we leverage sentential decision diagrams (SDDs) ~\cite{darwiche2011sdd} via  abstraction. SDDs are  tractable circuit representations that are at least as succinct as ordered binary decision diagrams (OBDDs) \cite{darwiche2011sdd}. Both of these  support  querying such as model counting (MC) and model enumeration (ME) in time linear in the size of the obtained circuit. (We use the term  querying  to mean both probabilistic conditional queries as well as weighted model counting because the latter simply corresponds to the case where the query is true.) Because of SDDs having such desirable properties, several papers have dealt with more involved issues, such as learning the structure from data directly~\cite{bekker2015tractable,liang2017learning} and thus learning the structure of the underlying graphical  model. 

In essence, our implementation regime uses SDDs as the underlying querying language for \wmi~in order to perform tractable and scalable probabilistic inference in hybrid domains. The regime neatly separates the model enumeration from the integration, which is demonstrated by allowing a choice of two integration schemes. The first is a provably efficient and exact integration approach for polynomial densities ~\cite{de2004effective,baldoni2011integrate,de2011software} and the second is an unmodified integration library available in the programming language platform (Python in our case). The results obtained are very promising with regards to the empirical behaviour: we perform competitively to the existing state-of-the-art \wmi~solver ~\cite{morettin2017efficient}. 
But perhaps most significantly, owing to the generic nature of our regime, we can scale the same approach to non-linear constraints, with possibly non-linear potentials.

\section{\uppercase{Background}}
\label{sec:background}
\textbf{Probabilistic Graphical Models.} Throughout this paper we will refer to Boolean and continuous random variables as $B_j$ and $X_i$ respectively for some finite $j>0,i>0$. Lower case letters, $b_j\in\{0,1\}$ and $x_i\in \mathbb{R}$, will represent the instantiations of these variables. Bold upper case letters will denote sets of variables and bold lower case letters will denote their instantiations.  We are broadly interested in probabilistic models, defined on $\mathbf{B}$ and $\mathbf{X}$. That is, let $(\mathbf{b},\mathbf{x}) = (b_1, b_2, \ldots ,b_m,x_1, x_2, \ldots ,x_n)$ be one element in the probability space $\{0,1\}^m \times \mathbb{R}^n$, denoting a particular assignment to the values in the respective domains.   A graphical model can then be used to describe dependencies between the variables and define a joint density function of those variables compactly.
The graphical model we will consider in this paper are Markov networks, which are undirected models. (Directed models can be considered too \cite{chavira2008probabilistic}, but are ignored for the sake of simplicity.) \smallskip 

{\bf Logical Background.} 
Propositional satisfiability (SAT)  is the of determining if a given formula in propositional logic can be satisfied by an assignment (, where a satisfying assignment has to be provided as proof for a formula being satisfiable). An instance of satisfiability modulo theory (SMT) ~\cite{biere2009handbook} is a generalization of classical SAT in allowing first-order formulas with respect to some decidable background theory. For example, 
$\mathcal{LRA}$ is understood here as  quantifier-free linear arithmetic formulas over the reals and the corresponding background theory is the fragment of first-order logic over the signature $(0,1,+,\leq)$, restricting the interpretation of these symbols to standard arithmetic.

In this work, we will consider two different background theories: quantifier-free linear ($\mathcal{LRA}$) and non-linear ($\mathcal{NRA}$) arithmetic over the reals.
A problem instance (input) to our $\wmi$ solver is then a formula with respect to one of those background theories in combination with propositional logic for which satisfaction is defined in an obvious way ~\cite{Barrett2009}. Such an instance is referred to as a \emph{hybrid knowledge base} (HKB). \smallskip

{\bf Weighted Model Counting.}
\label{sec:WMC}
Weighed model counting ($\wmc$) ~\cite{chavira2008probabilistic} is a strict generalization of model counting ~\cite{biere2009handbook}. In $\wmc$, each model of a given propositional knowledge base (PKB) $\Gamma$ has an associated weight and we are interested in computing the sum of the weights that correspond to models that satisfy $\Gamma$. (As is convention, the underlying propositional language and propositional letters are left implicit. We often refer to the set of literals \( \mathcal{L} \) to mean the set of all propositional atoms as well as their negations constructed from the propositions mentioned in $\Gamma$.)

In order to create an instance of the $\wmc$ problem given a PKB $\Gamma$ and literals $\mathcal{L}$, we define  a weight function $\mathit{wf}: \mathcal{L} \rightarrow \mathbb{R}^{\geq 0}$ mapping the literals to non-negative, numeric weights. We can then use the literals of a given model $m$ to define the weight of that model as well as the weighted model count as follows:

\begin{defn} 
\label{def:WMC}
Given a PKB $\Gamma$ over literals $\mathcal{L}$ (constructed from Boolean variables $\mathbf{B}$) and weight function $\mathit{wf}: \mathcal{L} \rightarrow \mathbb{R}^{\geq 0}$, we define the weight of a model as:
\begin{equation}
    \weight(m,\mathit{wf}) = \prod_{l\in m} \mathit{wf}(l)
\end{equation}{}
\noindent Further we define the weighted model count ($\wmc$) as:
\begin{equation}
    \wmc(\Gamma,\mathit{wf}) = \sum_{m \models \Gamma} \weight(m,\mathit{wf})
\end{equation}{}
\end{defn}
It can be shown that $\wmc$ can be used to calculate probabilities of a given graphical model $\mathcal{N}$ by means of a suitable encoding  \cite{chavira2008probabilistic}. In particular, conditional probabilities can be calculated using: $Pr_{\mathcal{N}}(q|\mathbf{e}) = \frac{\wmc(\Gamma \wedge q \wedge \mathbf{e},\mathit{wf})}{\wmc(\Gamma \wedge \mathbf{e},\mathit{wf})}$ for some evidence $\mathbf{e}$ and query $q$, where $\mathbf{e}, q$ are PKBs as well, defined from $\mathbf{B}$. \smallskip

{\bf Weighted Model Integration.} 
\label{sec:WMI}
While $\wmc$ is very powerful as an inference tool, it suffers from the inherent limitation of only admitting inference in discrete probability distributions. This is due to its underlying theory in enumerating all models (or expanding the complete network polynomial), which is exponential in the number of variables, but still finite and countable in the discrete case. For the continuous case, we need to find a language to reason about the uncountable event spaces, as well as represent density functions. $\wmi$  \cite{belle2015probabilistic} was proposed as a strict generalization of $\wmc$ for hybrid domains, with the  idea of annotating a  SMT theory with polynomial weights.

\begin{defn}~\cite{belle2015probabilistic}
\label{def:WMI}
Suppose $\Delta$ is a HKB over Boolean and real variables $\mathbf{B}$ and $\mathbf{X}$, and literals $\mathcal{L}$. Suppose $\mathit{wf}: \mathcal{L} \rightarrow \mathit{EXPR}(\mathbf{X})$, where $\mathit{EXPR}(\mathbf{X})$ are expressions over $\mathbf{X}$. Then we define \wmi~as:
\begin{equation}
    \wmi(\Delta,\mathit{wf}) = \sum_{m \models \Delta^-} \vol(m,\mathit{wf})
\end{equation}\noindent where:
\begin{equation}
    \vol(m,\mathit{wf}) = \int_{\{l^+ : l \in m\}} \weight(m,\mathit{wf}) d\mathbf{X}
\end{equation}{}
\noindent and $\weight$ is defined as described in Def~\ref{def:WMC}. 
\end{defn}
Intuitively the $\wmi$ of an SMT theory $\Delta$ is defined in terms of the models of its propositional abstraction $\Delta^-$. For each such model we compute its volume, that is, we integrate the $\weight$-values of the literals that are true in the model. The interval of the integral is defined in terms of the refinement of the literals. The weight function $\mathit{wf}$ is to be seen as mapping an expression $e$ to its density function, which is usually another expression mentioning the variables appearing in $e$. Conditional probabilities can be calculated as before. \smallskip 

{\bf Sentential Decision Diagram.}
Sentential decision diagrams (SDDs) were first introduced in ~\cite{darwiche2011sdd} and are graphical representations of propositional knowledge bases. SDDs are shown to be a strict subset of deterministic decomposable negation normal form (d-DNNF), a popular representation for probabilistic reasoning applications ~\cite{chavira2008probabilistic} due to their desirable properties. Decomposability and determinism ensure tractable probabilistic (and logical) inference, as they enable MAP queries in Markov networks. SDDs however satisfy two even stronger properties found in ordered binary decision diagrams (OBDD), namely structured decomposability and strong determinism. Indeed,~\cite{darwiche2011sdd} showed that they are strict supersets of OBDDs as well, inheriting their key properties: canonicity and a polynomial time support for Boolean combination. Finally SDD's also come with an upper bound on their size in terms of tree-width.
In the interest of space, we will not be able to discuss SDD properties in detail. However, we refer the reader to the original paper~\cite{darwiche2011sdd} for an in-depth study of SDDs and the central results of SDDs that we appeal to.

\section{\uppercase{Method}}
\label{sec:sharpNAR_Pipeline}

\noindent Over the past few years there have been several papers on exact probabilistic inference~\cite{morettin2017efficient,sanner2012symbolic,kolb2018efficient} using the formulation of \wmi. What we propose in this section is a novel formulation of doing weighted model integration by using SDDs as the underlying model counting, enumeration and  querying  language. Here predicate abstraction and knowledge compilation enable us to compile the abstracted PKB into an SDD, which has the desirable property of a fully parallelisable polytime model-enumeration algorithm. Recall that  polytime here refers to the complexity of the algorithm with respect to the size of the tree (SDD)~\cite{darwiche2002knowledge}.

In practice, computing the probability of a given query for some evidence consists of calculating the \wmi~of two separate but related HKBs. That is, we have to compute the \wmi~of a given HKB $\Delta$ conjoined with some evidence $\mathbf{e}$ and the query $q$, dividing it by the \wmi~of $\Delta$ conjoined with the evidence $\mathbf{e}$. This formulation introduced by~\cite{belle2015probabilistic} and explained in more detail in Section~\ref{sec:WMI}, can be written as:
\begin{equation}
    Pr_{\Delta}(q|\mathbf{e}) = \frac{\wmi(\Delta \wedge \mathbf{e} \wedge q)}{\wmi(\Delta \wedge \mathbf{e})}
\end{equation}{}

We will give a quick overview of the whole pipeline for computing the \wmi~value of a given KB, before discussing in detail the individual computational steps.

\subsection{\wmi-SDD: The Pipeline}
\begin{figure}
\centerline{\includegraphics[width=.37\textwidth]{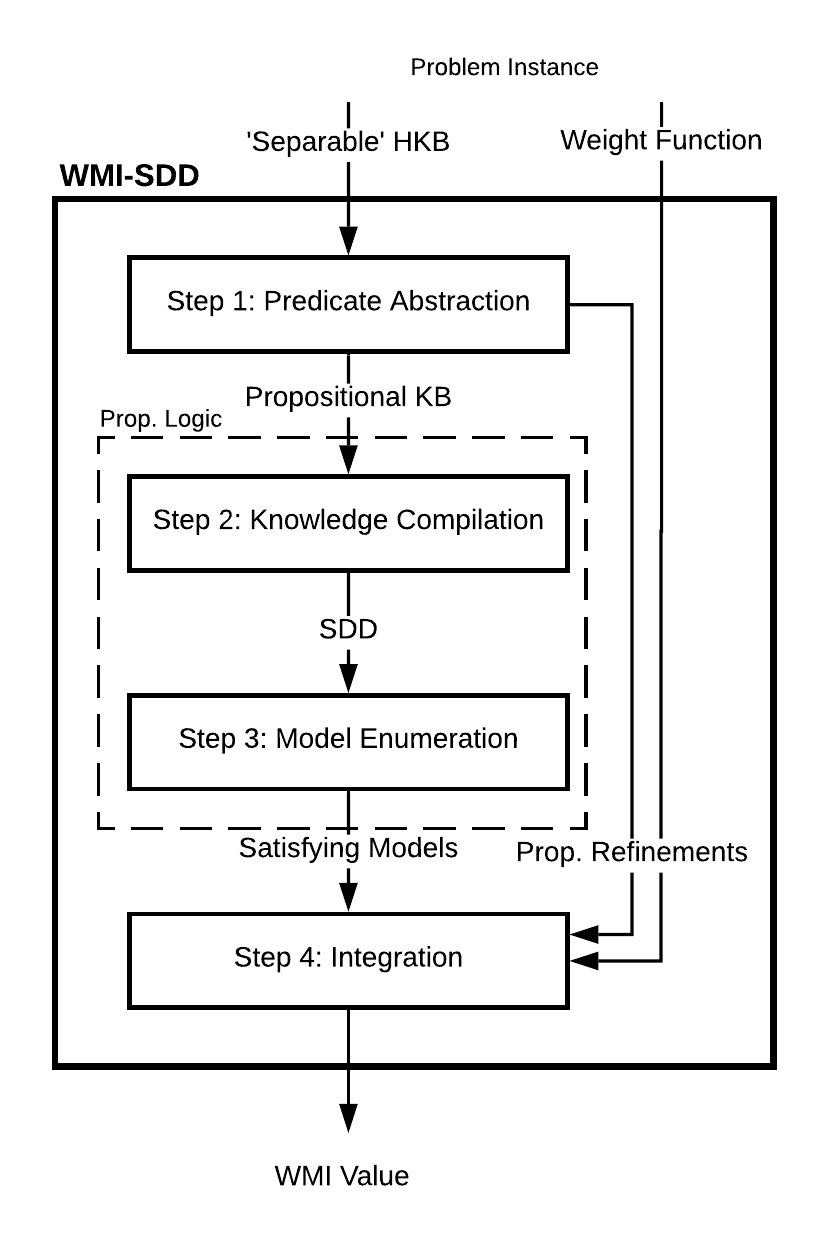}}
\caption{Pictorial depiction of the proposed pipeline for \wmi.}
  \label{fig:pipeline}
\end{figure}

As a basis for performing probabilistic inference, we first have to be able to calculate the \wmi~of a given HKB $\Delta$ with corresponding weight function $\mathit{wf}$. As we are interested in doing so by using SDDs as a query language, the \wmi~breaks down into a sequence of sub-computations depicted as the \wmi-SDD pipeline in Figure~\ref{fig:pipeline}.

{\bf Input/Outputs of the pipeline}
\label{sec:inputoutput}
The input of the pipeline is composed of two things: the HKB with respect to some background theory (eg. $\mathcal{LRA},\mathcal{NRA}$) on the one hand and the weight function on the other. Here, atoms are defined as usual for the respective language ~\cite{Barrett2009} and can be understood as functions that cannot be broken down further into a conjunction, disjunction or negation of smaller expressions.
This means that a HKB of the form $((X_1 < 3) \wedge (X_1 > 1))$ should be abstracted as $(B_1 \wedge B_2)$ with $B_1^{+} = (X_1 < 3)$ and $B_2^{+} = (X_1 > 1)$, rather than $B_0$ with $B_0^{+} = (X_1 < 3) \wedge (X_1 > 1)$.

The first step is to arrange atoms in a form that we call \lq separable\rq. The corresponding background theory determines whether a correct rewriting of formulas is possible to satisfy this condition: 

\begin{defn}
A given HKB $\Delta$ satisfies the condition \textbf{separable} if every atom within the formula can be rewritten in one of the following forms:  $X_1 < d(A)$, $d(A) < X_1$, $X_1 \leq d(A)$, $d(A) \leq X_1$ \text{ or } $d(A) \leq X_1 \wedge X_1 \leq d(A)$ where $d(A)$ is any term over $A \subseteq \vars - \{X_1\}$,  with  $\vars$ being the set of all variables (Boolean and continuous) that appear in the atom. That is, by construction, $X_1 \notin A$ for any given variable $X_1 \in \vars$. Such a variable $X_1$ is then called the \textbf{leading variable} (\textbf{leadVar}). 
\end{defn}

For some background theories, this conversion is immediate. In a $\mathcal{LRA}$ formula $\Delta_{\mathcal{LRA}}$, any given atom can be rewritten as an inequality or equality where we have a single variable on one side and a linear function on the other side, such as $(X_1 < 3 + X_2)$. But this is not a given for HKBs with background theory $\mathcal{NRA}$. For example, $(3 < 2*X_1 + X_2^2)$ can be rewritten as $(X_1< 3/2 - 1/2 * X_2^2)$ for $X_1$ and therefore satisfies the condition. However, the atom $(3 < X_1^4 - 3*X_1^2)$ cannot be rewritten in a similar manner and thus does not satisfy the condition.

Considering the motivation of performing probabilistic inference, where we deal with evidence and queries in addition to an HKBs, as discussed in Section~\ref{sec:WMI}, we note that all elements of $\{\Delta,q,\mathbf{e}\}$  have to fulfil the separability condition. As queries and evidence are applied by means of a logical connective with the HKB, they should generally be thought of as HKBs themselves.

The weight function $\mathit{wf}$, on the other hand, is only restricted by the condition that the term $\weight(m, \mathit{wf})$ must be integratable for any given model $m$. As long as this condition is met, we can accept any arbitrary function over the variables (Boolean and continuous) of the KB.

\subsection{Step 1: Predicate Abstraction} 
\label{sec:predicateAbstraction}

The aim of this step in the \wmi~framework is twofold. On the one hand, it is given an HKB ($\Delta$) and is tasked to produce a PKB ($\Delta^-$) and the corresponding mapping from propositional variables to continuous refinements, utilizing propositional abstraction. On the other hand, this part of the framework also rearranges the continuous refinements such that a single variable is separated from the rest of the equation to one side of the inequality/equality.

On a conceptual level, the predicate abstraction closely follows the theoretical formulation introduced in \cite{belle2015probabilistic}. The HKB is recursively traversed and every encountered atom is replaced with a propositional variable, while the logical structure (connectives and parentheses) of the KB is preserved.

We make use of the imposed \textit{separable} property to rewrite the individual refinements into bounds for a given variable. These bounds can easily be negated and will be used at a later stage to construct the intervals of integration for a given model. Now the process of rewriting a single atom corresponds to symbolically solving an equation for one variable and it is implemented as an arithmetic solver. The variable we choose to isolate from the rest of the equation (that is, the leading variable), is determined by a variable order, that in turn enforces the order of integration in a later stage of the pipeline. 
For example, assume that the chosen variable order is the usual alphabetical one over the variable names. Then predicates are rewritten such that from all variables referenced in the atom, the one highest up in the variable order is chosen as the leading variable and separated from the rest of the equation, resulting in a bound for the given variable. This ensures that for any predicate the bound for the leading variable does not reference any variable that precedes it alphabetically, which in turn ensures that the integral to be computed is defined and will result in natural number representing the volume. 

\begin{exmp}
\label{exmp:1}
To illustrate this with an example, consider the HKB $\Delta$: $\Delta = (B_0 \wedge (X_1 < 3) \wedge (0 < X_1 + X_2)) \vee (X_2 < 3 \wedge X_2 > 0)$.
After abstraction we are given the PKB $\Delta^- = $ ($B_0 \wedge B_1 \wedge B_2) \vee (B_3 \wedge B_4)$ where the abstracted variables correspond to the following atoms: $B_1^+ = ( X_1 < 3)$, $B_2^+ = (0 < X_1 + X_2)$, $B_3^+ = (X_2 < 3)$ and $B_4^+ = (X_2 > 0)$. 
As mentioned above, we construct the order of the continuous variable alphabetically, resulting in $\{1: X_1, 2:X_2\}$ for the proposed example. Once the order has been constructed we can rewrite each predicate as a bound for the variable appearing first in the order: $B_1 = (X_1 < 3)$, $B_2 = (-1 * X_2 < X_1)$, $B_3 = ( X_2 < 3)$ and $B_4 = (0 < X_2)$.  This ensures that the integral $\int \int \mathit{wf}(X_1,X_2) dX_1dX_2$ computes a  number for every possible model of the KB. Considering for example the model  $[B_0, B_1, B_2, B_3, B_4]$, the bounds of the integral would be as follows: $\int_{0}^3 \int_{-X_2}^3 \mathit{wf}(X_1,X_2) dX_1dX_2$ and yields a  number.
\end{exmp}

In the case of non-linear refinements, the step of rearranging the variable could give rise to new propositions, that in turn have to be added to the PKB. Consider, for example that the predicate $B$  with the refinement:  $B^+ = (4 < X_1 * X_2)$ should be rewritten for the variable $X_1$ as the leading one. Now as the variable $X_2$ might be negative or zero, we are unable to simply divide both sides by $X_2$ but rather have to split up the equation in the following way: $ B^+_{new} = (((X_2 > 0) \to (4/X_2 < X_1)) \wedge ((X_2 < 0) \to (4/X_2 > X_1)) \wedge ((X_2 = 0) \to False)) $ which can be further abstracted as: $B^+_{new} = ((B_1 \to B_2) \wedge (B_3 \to B_4) \wedge ((\neg B_1 \wedge \neg B_3) \to False))$. Once created, we can replace $B$ with its Boolean function refinement in the PKB and add all the new predicates $(B_1, B_2, B_3, B_4) $ to our list of propositions.

\subsection{Step 2: Knowledge Compilation}
In this step of our pipeline, the PKB constructed in the previous step is compiled into a canonical SDD. In practice, we first convert the PKB to CNF before passing it to the SDD library.\footnote{\url{http://reasoning.cs.ucla.edu/sdd/}} The library has a number of optimizations in place, including dynamic minimization \cite{choi2013dynamic}. However, the algorithm is still constrained by the asymptotically exponential nature of the problem. In addition, it requires the given PKB to be in CNF or DNF format.
Once the SDD is created, it is imported back into our internal data structure, which is designed for retrieving all satisfying models of a given SDD.

\subsection{Step 3: Model Enumeration}

Retrieving all satisfying models of a given PKB is a crucial part of the $\wmi$ formulation and we now focus on this step in our pipeline. In essence, we make use of knowledge compilation to compile the given PKB into a data structure, which allows us to enumerate all satisfying models in polynomial time with respect to the size of the tree. As discussed in the background section, SDDs are our data structures of choice and their properties, including canonicity, make them an appealing choice for our pipeline. 

The algorithm we developed for retrieving the satisfying models makes full use of the structural properties of SDDs. By recursively traversing the tree bottom-up, models are created for each node in the SDD with respect to the vtree node it represents. Those models are then passed upwards in the tree where they are combined with other branches. This is possible due to the structured decomposability property of the SDD data structure. It should also be noted at this point that parallelisation of the algorithm is possible as well due to SDDs decomposability properties. This is a highly desirable attribute when it comes to scaling to very large theories. 

\subsection{Step 4: Integration}

The workload of this part of the  framework is to compute the \textit{volume} (\vol) (as introduced in Def~\ref{def:WMI}) for every satisfying model that was found in the previous step. That volume for a given model of the PKB is computed by integrating the weight function ($\mathit{wf}$) over the literals true at the model, where the bound of the integral corresponds to the refinement and truth value of a given propositional variable within the model. All such volumes are then summed together and give the \wmi~value of the given HKB.

Computing a volume for a given model consists of two parts: firstly we have to combine the refinements of predicates appropriately, creating the bounds of integration before actually integrating over the $\mathit{wf}$ with respect to the variables and bounds. As discussed in the predicate abstraction and rewriting step, a given predicate (that has a refinement) consists of a leading variable and a bound for the variable. Combining the bounds into an interval is explained in Algorithm \ref{alg:interval}. 

\begin{algorithm}
\caption{Combining the intervals for a leadVar and model.}\label{algo1} \footnotesize
\begin{algorithmic}[1]
\Procedure{Combine}{\textit{leadVar, predicates, model}}
\State $\textit{interval} \gets (\textit{-inf}, \textit{inf})$
\For {$\textit{pred}\textbf{ in } predicates$}
\If{pred.leadVar $\ne$ leadVar}
\State continue
\EndIf
\If {$model[pred.idx] == \textit{false}$}
\State $newBound = negate(pred.bound)$
\Else
\State $newBound = pred.bound$
\EndIf
\State $interval = combine(interval, newBound)$
\EndFor
\State return \textit{interval}
\EndProcedure
\end{algorithmic}
\label{alg:interval}
\end{algorithm}

Here the function \textit{combine} combines intervals via intersections. For example, $combine((\textit{-inf},\textit{inf}), (\textit{-inf},X_1 < 3)) = (\textit{-inf}, min(\textit{inf}, 3)) = (\textit{-inf}, 3)$ and $combine((X_2 + X_3, \textit{inf}), (X_2/3 * X_2 < X_1, \textit{inf})) = (max(X_2 + X_3, X_3/3 * X_2), \textit{inf})$. This  procedure is done for every variable referenced in  $\mathit{wf}$, ensuring that we have a bound of integration for every such variable. 

Naturally, not all abstracted models have to be models of the original SMT theory. For example, suppose that a model makes both $X_0<5$ and $X_0>10$ true, abstracted as $B_1$ and $B_2$,  then the propositional abstraction erroneously retrieves a model where $[B_1,B_2, \ldots]$, and so the interval bounds would be $(10 < X_0 < 5)$. Clearly, then, the model should not be considered as a model for the SMT theory and is simply disregarded. Once all the real bounds of integration are defined for the given model, the next step before integrating is to enumerate all possible instantiations of Boolean variables referenced in the $\mathit{wf}$. The different integration problems are then hashed such that the system only has to compute the integration once, even if they appear multiple times.

When it comes to the implementation of this part of the framework, we used two different integration methods. We support the  integration module of  the scipy python package\footnote{\url{https://scipy.org/}} to compute the defined integral for a given $\mathit{wf}$, a set of intervals and the instantiations of Boolean variables. Using this package allowed us to formalize the method as described above and perform inference in non-linear domains. However, this formulation is not exact and suffers from a slow runtime. For this reason, we also implemented the pipeline using \textit{latte},\footnote{\url{https://www.math.ucdavis.edu/~latte/}} an exact integration software that is particularly well-suited for piecewise polynomial density approximations.

\section{\uppercase{Empirical Evaluation}}

\noindent Here, we evaluate the proposed framework on the time it needs to compute the \wmi~of a given HKB and $\mathit{wf}$. It is a proof-of-concept system for \wmi~via  SDDs. To evaluate the framework, we randomly generate  problems, as described below and compare the time to the \wmi-PA framework developed in  \cite{morettin2017efficient}.\footnote{We were unable to compare the performance with the framework developed in \cite{kolb2018efficient} owing  to compatibility issues in the experimental setup. Since it is reported to perform comparably to \cite{morettin2017efficient}, all comparisons made in this paper are in reference to the pipeline developed in \cite{morettin2017efficient}.}

\subsection{Problem Set Generator}
A problem is generated based on 3 factors: the number of variables, the number of clauses and the percentage of real variables. 

When generating a new Boolean atom, we simply return a Boolean variable with the given ID, whereas generating a real-valued atom is more intricate and depends on the kind of HKB we are generating (i.e., $\mathcal{LRA}$ vs $\mathcal{NRA}$). For both background theories we generate a constant interval for a given variable ID with probability $0.5$ (e.g., $345 < X_3 < 789$ for variable ID $3$). Otherwise, we pick two random subsets of all other real variables $\mathbf{X}_L, \mathbf{X}_U \subset \vars_{\mathit{Real}}$ for the upper and lower bound respectively. Now if we are generating an HKB with respect to the background theory $\mathcal{LRA}$, we sum all variables in the upper as well as the lower bound, to create a linear function as the upper and lower bound for the variable. Similarly, when generating an HKB with respect to the background theory $\mathcal{NRA}$, we conjoin the variables of a given set ($\mathbf{X}_L, \mathbf{X}_U$) by multiplication rather than by addition. Finally, when creating such an interval we additionally add a constant interval for the same variable ID to make sure our integration is definite and evaluates to a real number.

In order to evaluate our framework, we let the number of variables ($\mathit{nbVars}$) range from 2 to 28, where the number of clauses we tested is $\mathit{nbVars} * 0.7$, $\mathit{nbVars}$ and $\mathit{nbVars} * 1.5$ for a given value of $\mathit{nbVars}$. Now for each variable clause pair, we generate two problem instances where the percentage of continuous variables is set to 50\% to account for the randomness of the generator. Thus for each number of variables, we generate six different problems, which are then averaged to compute a final runtime.

\section{\uppercase{Results}}

\begin{figure}
  \centering 
  \includegraphics[width=.4\textwidth]{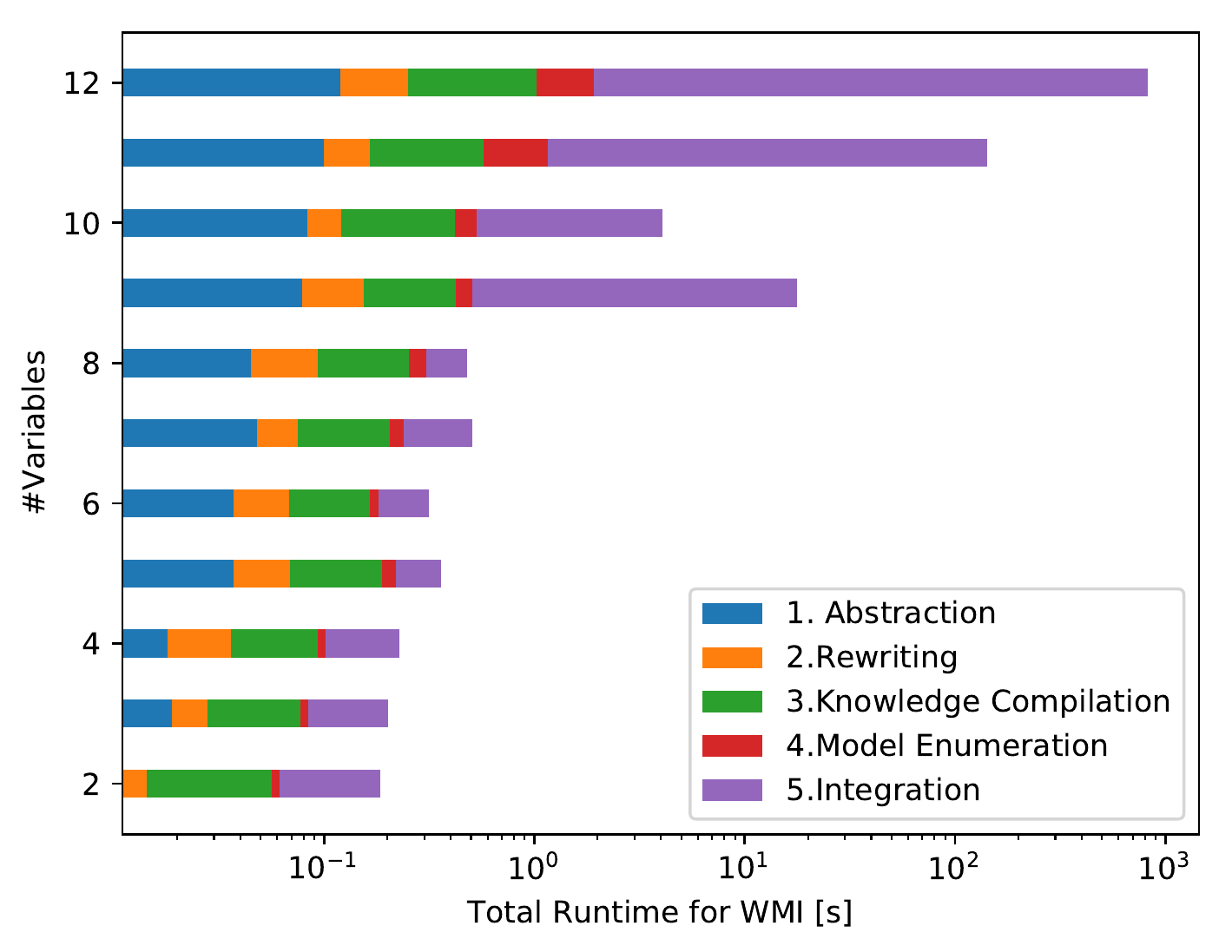}
  \caption{Runtime analysis of $\wmi$-SDD framework for non-linear HKBs.}
  \label{fig:runtime_analysis}
\end{figure}

First, we discuss the performance of our framework on non-linear hybrid domains. As part of this experiment the generated HKB consists of non-linear atoms which are products of variables (e.g. $X_1 * X_2 * -4 * X_3 < X_4 < X_1 * 27 *X_5$). Figure~\ref{fig:runtime_analysis} plots the average time spent in each computational step for all problems that have the same number of variables. Here we see that the overall time increases with the number of variables as expected. While most of the steps have a rather small impact on the overall computational time, the integration step has by far the greatest cost. This is in part due to the Scipy integration method, which was used for these benchmarks, as it can cope with non-linear bounds but is not as efficient as the latte integration package. Finally, we want to point out the surprisingly small cost of compiling the PKB into an SDD, which reinforces our decision to use knowledge compilation.

\begin{figure}
  \centering 
  \includegraphics[width=.4\textwidth]{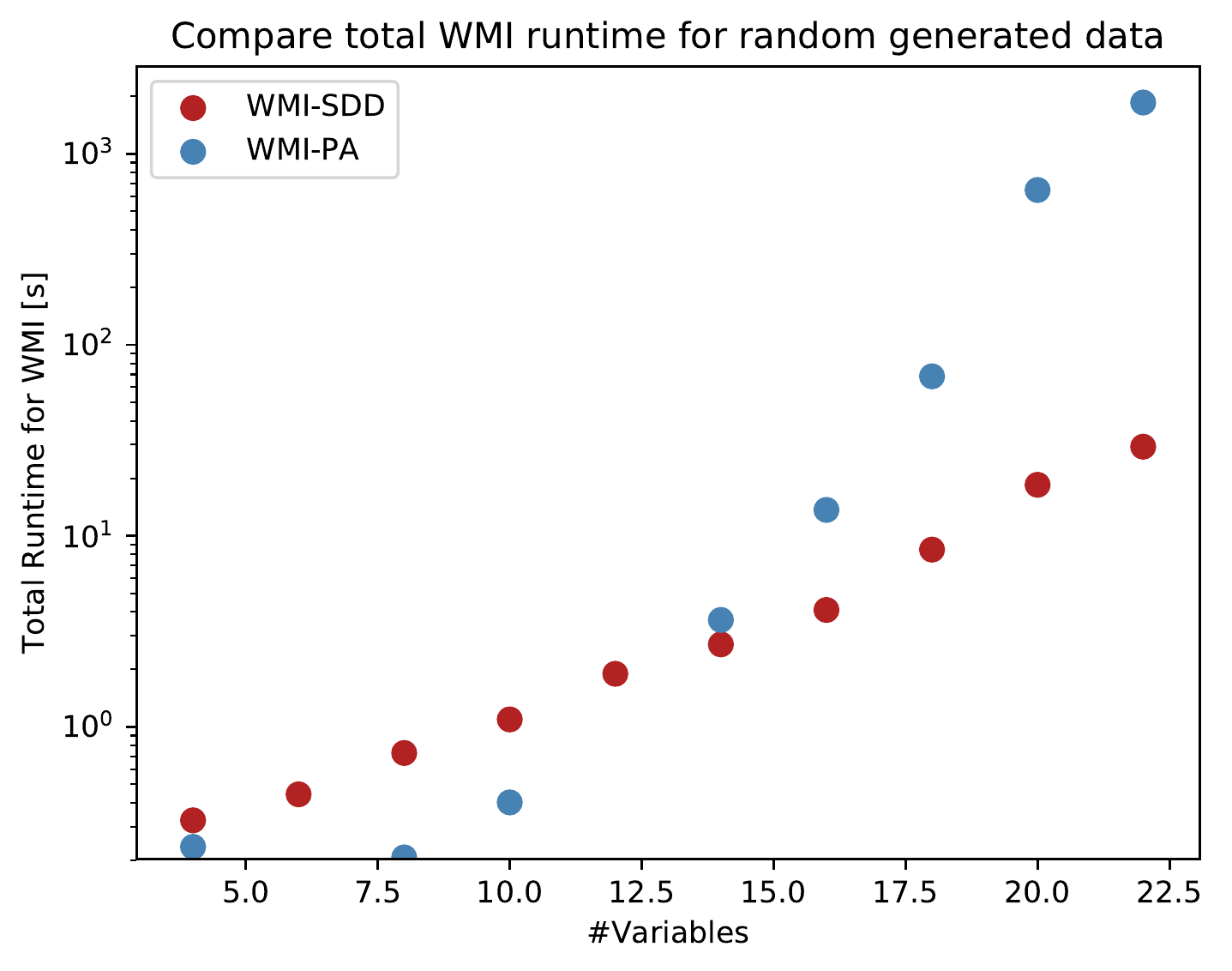}
  \caption{Total runtime comparison \wmi-SDD vs \wmi-PA for linear HKBs.}
  \label{fig:runtime_compare}
\end{figure}

Next, we  discuss the performance of the \wmi-SDD framework on linear HKBs against one of the current state-of-the-art \wmi~solver, the \wmi-PA framework ~\cite{morettin2017efficient}.  The results are plotted  in Figure~\ref{fig:runtime_compare}. The results  demonstrate the overall impact of using knowledge compilation as part of the framework. While the additional step of compiling the abstracted PKB into an SDD results in longer computational time for small problem instances, the trade-off shows its advantage as we increase the number of variables. Considering the logarithmic scale of the y-axis, the difference between the two algorithms becomes quite substantial as the number of variables exceeds 20. By extension, we believe the \wmi-SDD framework shows tremendous promise for scaling \wmi~to large domains in the future.

Before concluding this section, we remark that readers familiar with propositional model counters are likely to be surprised by the total variable size being less than 50 in our experiments and other \wmi~solvers \cite{morettin2017efficient}. Contrast this with SDD evaluations that scale to hundreds of propositional variables  \cite{darwiche2011sdd,choi2013dynamic}. The main bottleneck here is symbolic integration, even if in isolation solvers such as \textit{latte} come with strong polynomial time bounds~\cite{baldoni2011integrate}. This is because integration has been performed for each model, and so with $n$ variables and a knowledge base of the form $(a_ 1 < X_ 1 < b_ 1) \lor \ldots \lor (a_ n < X_ n < b_ n)$, where $a_i,b_j \in \mathbb{R}$, there are $ 2^n \times n$ integration computations in the worst case. That is, there are $ 2^n$ models on abstraction, and in each model, we will have $n$ integration variables. 

There are a number of possible ways to address that concern. First, a general solution is to simply focus on piecewise constant potentials, in which case, after abstraction, \wmi~over an HKB immediately reduces to a $\wmc$ task over the corresponding PKB. Second, parallelisation can be enabled. For example, we can decompose a CNF formula into \textit{components}, which are CNF formulas themselves, the idea being that components do not share variables ~\cite{series/faia/GomesSS09}. In this case, the model count of a formula $F$, written $\#F$ with $n$ components $C_1, \ldots, C_n$ would be $\#C_1 \times \cdots \times \#C_n$. This is explored for the interval fragment in ~\cite{belle2016component}. Third, one can keep a dictionary of partial computations of the integration (that is, cache the computed integrals), and apply these values where applicable.

While we do not explore such possibilities in this article, we feel the ability of SDDs to scale as well as its ability to enable parallelisation can be seen as additional justifications for our approach. We also suspect that it should be fairly straightforward to implement such choices given the modular way our solver is realized.

\section{\uppercase{Conclusion}}
\label{sec:conclusion}

In this paper, we introduced a novel way of performing \wmi~by leveraging efficient predicate abstraction and knowledge compilation. Using SDDs to represent the abstracted HKBs enabled us to make full use of the structural properties of SDD and devise an efficient algorithm for retrieving all satisfying models. The evaluations demonstrate the competitiveness of our framework and reinforce our hypothesis that knowledge compilation is worth considering even in continuous domains. We were also able to deal with a specific class of separable non-linear constraints.

In the future, we would like to better explore how the integration bottleneck can be addressed, possibly by caching sub-integration computations. In independent recent efforts, ~\cite{zuidberg2019exact,kolb2019exploit} also investigate the use of SDDs for performing \wmi. In particular, ~\cite{kolb2019exploit} consider a different type of mapping between \wmi~and SDDs but do not consider non-linear domains, whereas  ~\cite{zuidberg2019exact} allow for standard density functions such as Gaussians by appealing to algebraic model counting~\cite{KimmigJAL16}. Performing additional comparisons and seeing how these ideas could be incorporated in our framework might be an interesting direction for the future.

\section*{\uppercase{Acknowledgements}}

\noindent Anton Fuxjaeger was supported by the Engineering and Physical Sciences Research Council (EPSRC) Centre for Doctoral Training in Pervasive Parallelism (grant EP/L01503X/1) at the School of Informatics, University of Edinburgh. Vaishak Belle was supported by a Royal Society University Research Fellowship. We would also like to thank our reviewers for their helpful suggestions.

\bibliographystyle{apalike}
{\small
\bibliography{main}}

\begin{thebibliography}{}

\bibitem[{Albarghouthi} et~al., 2017]{albarghouthi2017quantifying}
{Albarghouthi}, A., {D'Antoni}, L., {Drews}, S., and {Nori}, A. (2017).
\newblock {Quantifying Program Bias}.
\newblock {\em arXiv e-prints}, page arXiv:1702.05437.

\bibitem[Bahar et~al., 1997]{bahar1997algebric}
Bahar, R.~I., Frohm, E.~A., Gaona, C.~M., Hachtel, G.~D., Macii, E., Pardo, A.,
  and Somenzi, F. (1997).
\newblock Algebric decision diagrams and their applications.
\newblock {\em Formal methods in system design}, 10(2-3):171--206.

\bibitem[Baldoni et~al., 2011]{baldoni2011integrate}
Baldoni, V., Berline, N., De~Loera, J., K{\"o}ppe, M., and Vergne, M. (2011).
\newblock How to integrate a polynomial over a simplex.
\newblock {\em Mathematics of Computation}, 80(273):297--325.

\bibitem[Barrett et~al., 2009]{Barrett2009}
Barrett, C.~W., Sebastiani, R., Seshia, S.~A., and Tinelli, C. (2009).
\newblock Satisfiability modulo theories.
\newblock In \cite{biere2009handbook}, pages 825--885.

\bibitem[Bekker et~al., 2015]{bekker2015tractable}
Bekker, J., Davis, J., Choi, A., Darwiche, A., and Van~den Broeck, G. (2015).
\newblock Tractable learning for complex probability queries.
\newblock In {\em Advances in Neural Information Processing Systems}, pages
  2242--2250.

\bibitem[Belle et~al., 2015]{belle2015probabilistic}
Belle, V., Passerini, A., and Van~den Broeck, G. (2015).
\newblock Probabilistic inference in hybrid domains by weighted model
  integration.
\newblock In {\em Proceedings of 24th International Joint Conference on
  Artificial Intelligence (IJCAI)}, pages 2770--2776.

\bibitem[Belle et~al., 2016]{belle2016component}
Belle, V., Van~den Broeck, G., and Passerini, A. (2016).
\newblock Component caching in hybrid domains with piecewise polynomial
  densities.
\newblock In {\em Proceedings of the 30th Conference on Artificial Intelligence
  (AAAI)}.

\bibitem[Biere et~al., 2009]{biere2009handbook}
Biere, A., Biere, A., Heule, M., van Maaren, H., and Walsh, T. (2009).
\newblock {\em Handbook of Satisfiability: Volume 185 Frontiers in Artificial
  Intelligence and Applications}.
\newblock IOS Press, Amsterdam, The Netherlands, The Netherlands.

\bibitem[Chavira and Darwiche, 2008]{chavira2008probabilistic}
Chavira, M. and Darwiche, A. (2008).
\newblock On probabilistic inference by weighted model counting.
\newblock {\em Artificial Intelligence}, 172(6-7):772--799.

\bibitem[Chistikov et~al., 2017]{chistikov2017approximate}
Chistikov, D., Dimitrova, R., and Majumdar, R. (2017).
\newblock Approximate counting in smt and value estimation for probabilistic
  programs.
\newblock {\em Acta Informatica}, 54(8):729--764.

\bibitem[Choi and Darwiche, 2013]{choi2013dynamic}
Choi, A. and Darwiche, A. (2013).
\newblock Dynamic minimization of sentential decision diagrams.
\newblock In {\em AAAI}.

\bibitem[Choi et~al., 2013]{choi2013compiling}
Choi, A., Kisa, D., and Darwiche, A. (2013).
\newblock Compiling probabilistic graphical models using sentential decision
  diagrams.
\newblock In {\em European Conference on Symbolic and Quantitative Approaches
  to Reasoning and Uncertainty}, pages 121--132. Springer.

\bibitem[Darwiche, 2004]{darwiche2004new}
Darwiche, A. (2004).
\newblock New advances in compiling cnf to decomposable negation normal form.
\newblock In {\em Proceedings of the 16th European Conference on Artificial
  Intelligence}, pages 318--322. Citeseer.

\bibitem[Darwiche, 2011]{darwiche2011sdd}
Darwiche, A. (2011).
\newblock Sdd: A new canonical representation of propositional knowledge bases.
\newblock In {\em IJCAI Proceedings-International Joint Conference on
  Artificial Intelligence}, volume~22, page 819.

\bibitem[Darwiche and Marquis, 2002]{darwiche2002knowledge}
Darwiche, A. and Marquis, P. (2002).
\newblock A knowledge compilation map.
\newblock {\em Journal of Artificial Intelligence Research}, 17(1):229--264.

\bibitem[De~Loera et~al., 2011]{de2011software}
De~Loera, J., Dutra, B., Koeppe, M., Moreinis, S., Pinto, G., and Wu, J.
  (2011).
\newblock Software for exact integration of polynomials over polyhedra.
\newblock {\em arXiv preprint arXiv:1108.0117}.

\bibitem[De~Loera et~al., 2004]{de2004effective}
De~Loera, J.~A., Hemmecke, R., Tauzer, J., and Yoshida, R. (2004).
\newblock Effective lattice point counting in rational convex polytopes.
\newblock {\em Journal of symbolic computation}, 38(4):1273--1302.

\bibitem[Fierens et~al., 2015]{fierens2015inference}
Fierens, D., Van~den Broeck, G., Renkens, J., Shterionov, D., Gutmann, B.,
  Thon, I., Janssens, G., and De~Raedt, L. (2015).
\newblock Inference and learning in probabilistic logic programs using weighted
  boolean formulas.
\newblock {\em Theory and Practice of Logic Programming}, 15(3):358--401.

\bibitem[Gomes et~al., 2009]{series/faia/GomesSS09}
Gomes, C.~P., Sabharwal, A., and Selman, B. (2009).
\newblock Model counting.
\newblock In \cite{biere2009handbook}, pages 633--654.

\bibitem[Kimmig et~al., 2016]{KimmigJAL16}
Kimmig, A., Van~den Broeck, G., and De~Raedt, L. (2016).
\newblock Algebraic model counting.
\newblock {\em International Journal of Applied Logic}.

\bibitem[Kisa et~al., 2014]{kisa2014probabilistic}
Kisa, D., Van~den Broeck, G., Choi, A., and Darwiche, A. (2014).
\newblock Probabilistic sentential decision diagrams.
\newblock In {\em KR}.

\bibitem[Kolb et~al., 2018]{kolb2018efficient}
Kolb, S., Mladenov, M., Sanner, S., Belle, V., and Kersting, K. (2018).
\newblock Efficient symbolic integration for probabilistic inference.
\newblock In {\em IJCAI}, pages 5031--5037.

\bibitem[Kolb et~al., 2019]{kolb2019exploit}
Kolb, S., Zuidberg Dos~Martires, P.~M., and De~Raedt, L. (2019).
\newblock How to exploit structure while solving weighted model integration
  problems.
\newblock {\em UAI 2019 Proceedings}.

\bibitem[Liang et~al., 2017]{liang2017learning}
Liang, Y., Bekker, J., and Van~den Broeck, G. (2017).
\newblock Learning the structure of probabilistic sentential decision diagrams.
\newblock In {\em Proceedings of the 33rd Conference on Uncertainty in
  Artificial Intelligence (UAI)}.

\bibitem[Martires et~al., 2019]{zuidberg2019exact}
Martires, P., Dries, A., and De~Raedt, L. (2019).
\newblock Exact and approximate weighted model integration with probability
  density functions using knowledge compilation.
\newblock {\em Proceedings of the AAAI Conference on Artificial Intelligence},
  33:7825--7833.

\bibitem[Morettin et~al., 2017]{morettin2017efficient}
Morettin, P., Passerini, A., and Sebastiani, R. (2017).
\newblock Efficient weighted model integration via smt-based predicate
  abstraction.
\newblock In {\em Proceedings of the Twenty-Sixth International Joint
  Conference on Artificial Intelligence, {IJCAI-17}}, pages 720--728.

\bibitem[Muise et~al., 2012]{muise2012d}
Muise, C., McIlraith, S.~A., Beck, J.~C., and Hsu, E.~I. (2012).
\newblock D sharp: fast d-dnnf compilation with sharpsat.
\newblock In {\em Canadian Conference on Artificial Intelligence}, pages
  356--361. Springer.

\bibitem[Poon and Domingos, 2011]{poon2011sum}
Poon, H. and Domingos, P. (2011).
\newblock Sum-product networks: A new deep architecture.
\newblock In {\em Computer Vision Workshops (ICCV Workshops), 2011 IEEE
  International Conference on}, pages 689--690. IEEE.

\bibitem[Sang et~al., 2005]{sang2005performing}
Sang, T., Beame, P., and Kautz, H.~A. (2005).
\newblock Performing bayesian inference by weighted model counting.
\newblock In {\em AAAI}, volume~5, pages 475--481.

\bibitem[Sanner et~al., 2012]{sanner2012symbolic}
Sanner, S., Delgado, K., and Barros, L. (2012).
\newblock Symbolic dynamic programming for discrete and continuous state mdps.
\newblock {\em CoRR}, abs/1202.3762.

\bibitem[Shenoy and West, 2011]{shenoy2011inference}
Shenoy, P.~P. and West, J.~C. (2011).
\newblock Inference in hybrid bayesian networks using mixtures of polynomials.
\newblock {\em International Journal of Approximate Reasoning}, 52(5):641--657.

\bibitem[Suciu et~al., 2011]{suciu2011probabilistic}
Suciu, D., Olteanu, D., R{\'e}, C., and Koch, C. (2011).
\newblock Probabilistic databases.
\newblock {\em Synthesis lectures on data management}, 3(2):1--180.

\bibitem[Van~den Broeck and Darwiche, 2015]{van2015role}
Van~den Broeck, G. and Darwiche, A. (2015).
\newblock On the role of canonicity in knowledge compilation.
\newblock In {\em AAAI}, pages 1641--1648.

\end{thebibliography}

\end{document}